\documentclass{article}
\usepackage[final]{bdl_2018}

\usepackage[utf8]{inputenc} 
\usepackage[T1]{fontenc}    
\usepackage{hyperref}       
\usepackage{url}            
\usepackage{booktabs}       
\usepackage{amsfonts}       
\usepackage{nicefrac}       
\usepackage{microtype}      
\usepackage{amsmath}
\usepackage[pdftex]{graphicx}
\usepackage{subcaption}
\usepackage{comment}

\title{Informed MCMC with Bayesian Neural Networks\\for Facial Image Analysis}

\author{
  Adam Kortylewski\thanks{Equal contribution.}, Mario Wieser\footnotemark[1], Andreas Morel-Forster\footnotemark[1], Aleksander Wieczorek, \\
  \textbf{Sonali Parbhoo, Volker Roth, Thomas Vetter}\\
  Department of Mathematics and Computer Science\\
  University of Basel\\
}

\begin{document}

\maketitle
\section{Introduction}

\textbf{Motivation.}
Computer vision tasks are difficult because of the large variability in the data that is induced by changes in light, background, partial occlusion as well as the varying pose, texture and shape of objects. Generative approaches to computer vision allow us to overcome this difficulty by explicitly modeling the physical image formation process. Such models can produce images $R(y)$ using a deterministic rendering engine $R$ and a set of parameters $y$ that define the scene in terms of e.g. light sources and object properties. The analysis of an observed image $x$ is then performed via Bayesian inference of the posterior distribution $p(y | x)$.

\textbf{Problem.} This conceptually simple approach tends to fail in practice because of several difficulties stemming from sampling the posterior distribution \cite{jampani2015informed}: high-dimensionality and multi-modality of the posterior distribution $p(y|x)$ as well as expensive simulation of the rendering process. Sampling $p(y|x)$ is typically performed with a Markov Chain Monte Carlo algorithm, such as Metropolis-Hastings \cite{schonborn2017markov,jampani2015informed}. The general idea is to sequentially generate samples from the posterior distribution of $y$ by performing the following two steps:
\begin{enumerate}
	\item Generate a new point from the proposal distribution: $	y_{t+1} \sim Q(\cdot | y_t)$.
	\item Accept the new point with the acceptance probability:\\ $A(y_{t+1}, y_t) = \mbox{min}\left( 1, \frac{p(y_{t+1}) Q(y_{t}|y_{t+1})}{p(y_{t}) Q(y_{t+1}|y_{t})}\right)$.	
\end{enumerate}

The main difficulty of MCMC in a computer vision context is how to choose the proposal distribution accurately so that maxima of the posterior are explored early and the Markov chain quickly converges to a valid image interpretation. 

\textbf{Contribution.} In this work, we propose to use a Bayesian Neural Network for estimating an \textit{image dependent} proposal distribution $Q(\cdot|x)$. Compared to a standard Gaussian random walk proposal, this will accelerate the sampler in finding regions of the posterior with high value. In this way, we can significantly reduce the number of samples needed to perform facial image analysis.

\section{Methodology} 
\textbf{Generative Face Model.} In the context of facial image analysis, the 3D Morphable Model (3DMM) \cite{gerig2018morphable} is commonly used as prior for the 3D face geometry, color as well as the computer graphics parameters needed for the rendering process.
Sch\"onborn et al. \cite{schonborn2017markov} proposed using the
Metropolis Hastings algorithm to estimate the posterior
over the model parameters $y$:
\begin{equation}
    p(y|x) \sim p(x|y) p(y).
\end{equation}
The likelihood $p(x|y)$ measures the similarity between the
target image $x$ and the rendered image $R(y)$ assuming
pixel-wise independence. Given a posterior estimate, we can perform a
multitude of facial image analysis tasks, such as face recognition
\cite{blanz2003face}, 3D face reconstruction
\cite{schonborn2017markov} or face manipulation
\cite{walker2009portraits}.

\textbf{Informed Sampler.} A key component of an MCMC sampler is the proposal distribution $Q(y)$. In the context of computer vision,  $Q(y)$ needs to be carefully tuned in order to explore the posteriors maxima in a reasonable time. In order to overcome this limitation Jampani et al. \cite{jampani2015informed} propose to combine a local Gaussian random walk proposal $Q_L$ with an image dependent, global proposal distribution $Q_I$:
\begin{equation}
    y_{t+1} \sim \alpha Q_L(\cdot|y_t) + (1-\alpha) Q_I(\cdot|x).
\end{equation}
The global proposal distribution is estimated discriminatively based on
the input image. In \cite{jampani2015informed} the authors propose to use manually designed image
features and a kernel density estimate for estimating $Q_I(\cdot|x)$. We instead propose to learn this distribution from data using Bayesian Neural Networks.

\textbf{Bayesian Neural Networks.} A Bayesian Neural Network (BNN) estimates, in contrast to traditional Neural Networks, not only a point estimate but also the corresponding uncertainties. In \cite{NIPS2017_7141}, Kendall and Gal describe model (Epistemic) and data (Heteroscedastic Aleatoric) uncertainties to be crucial for computer vision tasks and introduce an approach to unify both uncertainties within a BNN. We build upon this approach and estimate our global distribution $Q_I(\cdot|x)$ with a BNN which is in turn used to inform the MCMC sampler. In doing so, we place a prior distribution over the neural network weights $W$ to capture the model uncertainty. We estimate the posterior distribution of $W$ during training using Bayesian inference given our training data $X=\{\ x_1, \dots, x_N \}$ and $Y\{\ y_1, \dots, y_N \}$:
\begin{equation}
p(W \mid X,Y) = \frac{p(X,Y \mid W) p(W)}{\int p(Y \mid X,W)p(W)dW}
\end{equation}
Subsequently, we formulate our data uncertainty in terms of a Gaussian likelihood $p(y \mid f^W(x)) =  \mathcal{N}(f^W(x),\,\sigma^{2})$ because the 3DMM parameters $Y$ are continuous values. Here, the mean is denoted as our model output $f^W(x)$ and $\sigma^2$ defines the corresponding variance. Having estimated both uncertainties, we combine these uncertainties as described in \cite{NIPS2017_7141} to obtain our informed proposal distribution $Q_I(\cdot|x)$.

\section{Experiments}
\textbf{Datasets and setup.} We train our BNN on synthetic data in similar to as proposed by Kim et al. \cite{kim2017inversefacenet}. We train an AlexNet \cite{krizhevsky2012imagenet} architecture using 300K synthetically generated face images with corresponding 3DMM parameters $\{X,Y\}$. The code and data used for our experiments will be made available \footnote{https://github.com/unibas-gravis/bnn-informed-face-sampler}. For testing, we use a sample of 150 face images from the CMU-Multipie face dataset \cite{gross2010multi}, sampled from Session-01 using the frontal and $30^\circ$ cameras.
    \begin{figure}[h]
    	\begin{subfigure}{0.18\linewidth}
    		\centering
    		\includegraphics[height=2.3cm]{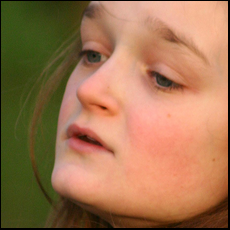}
    		\caption{}
    		\label{fig:bnn-input}
    	\end{subfigure}%
    	\begin{subfigure}{0.18\linewidth}
    		\centering
    		\includegraphics[height=2.3cm]{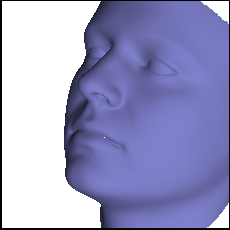}
    		\caption{}
    		\label{fig:bnn-mean}
    	\end{subfigure}
    	\begin{subfigure}{0.49\linewidth}
    		\centering
    		\includegraphics[height=2.3cm]{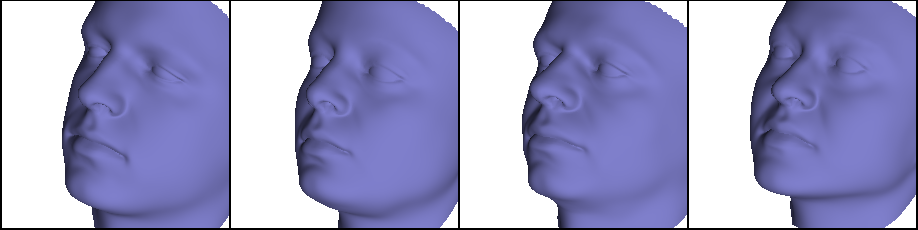}
    		\caption{}
    		\label{fig:bnn-samples}
    	\end{subfigure}
    	\caption{Uncertain 3D face reconstruction with BNNs. We illustrate the joint pose and shape distribution for simplicity. Note that our model predicts a joint distribution over all 3DMM parameters. (a) The test image from the AFLW dataset \cite{koestinger11a}. (b) The mean prediction of our BNN. (c) Samples from the joint prediction uncertainty in head pose and 3D shape. Notably, the variability in the head pose is low, whereas the remaining variability in the shape is comparably high (e.g. the nose region).}
    	\label{fig:bnn}
    \end{figure}
    
\textbf{Qualitative results.} Figure \ref{fig:bnn} illustrates the prediction uncertainty of our model given the test image in Figure \ref{fig:bnn-input}. Note that the mean prediction (Figure \ref{fig:bnn-mean}) has a correct head pose and a similar 3D face shape as the face in the test image. The prediction uncertainty of our model (Figure \ref{fig:bnn-samples}) is visualized by sampling from the normal distribution defined by the mean prediction and the joint uncertainty estimated as described in the previous section. Note the remaining variability in the head pose is low, whereas it is large in the shape, e.g. in the nose region. This observation is reasonable as the 3D head pose can in principle be estimated from the 2D spatial configuration of a few facial features, whereas the estimation of the 3D face geometry from a single monocular image is ill-posed.

\begin{figure}[h]
	\begin{subfigure}{0.32\linewidth}
		\centering
		\includegraphics[height=3.5cm]{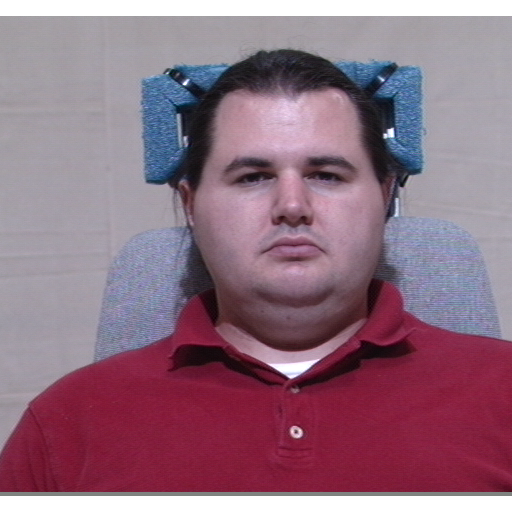}
		\caption{}
		\label{fig:inf-frontal}
	\end{subfigure}%
	\begin{subfigure}{0.3\linewidth}
		\centering
		\includegraphics[height=3.5cm]{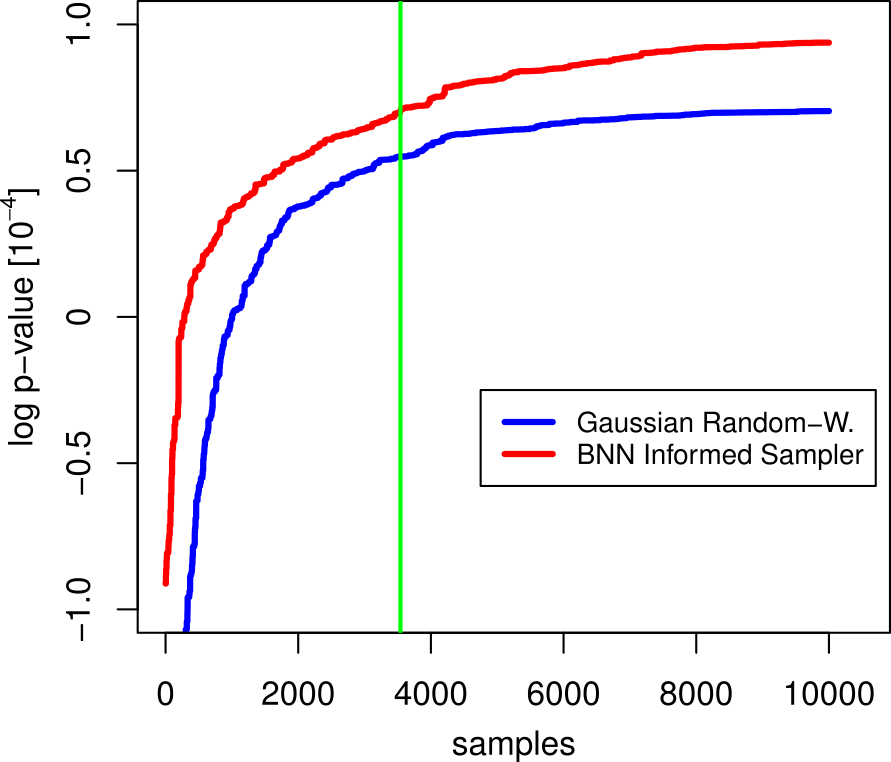}
		\caption{}
		\label{fig:inf-posterior}
	\end{subfigure}	
	\begin{subfigure}{0.2\linewidth}
		\centering
		\includegraphics[height=3.5cm]{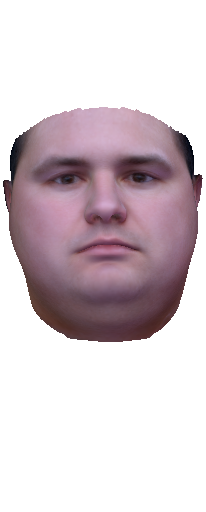}
		\caption{}
		\label{fig:inf-informed}
	\end{subfigure}%
	\begin{subfigure}{0.2\linewidth}
		\centering
		\includegraphics[height=3.5cm]{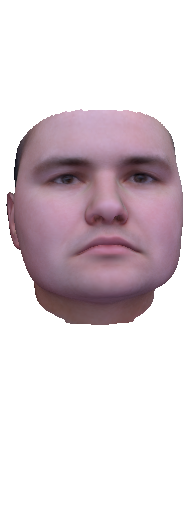}
		\caption{}
		\label{fig:inf-uninformed}
	\end{subfigure}%
	\caption{Comparison of uninformed and BNN-informed MCMC. (a) The test image. (b) Maximal observed posterior over 10K samples for both MCMC approaches. Our informed sampler (red curve) explores high posterior values faster than the uninformed sampler (blue curve). It also reaches the same maximal posterior value after already about 3500 samples (green vertical line). Therefore it can obtain a better image interpretation (c) compared to an uninformed sampler (d) within a fixed frame of 10K samples.}
	\label{fig:inf}
\end{figure}

\textbf{Quantitative results.} We integrate the uncertain prediction of our BNN into Markov Chain Monte Carlo (MCMC) sampling as an informed proposal distribution $Q_I(\cdot|x)$. In Figure \ref{fig:inf} we compare our BNN-informed sampling to one with an uninformed block-wise Gaussian proposal distribution when applied to the test image in Figure \ref{fig:inf-frontal}. When plotting the maximal unnormalized posterior over runs of 10000 samples we can observe that the proposed BNN-informed MCMC (red curve) explores samples with high posterior values earlier compared to the uninformed sampler (blue curve). From the plot, we can also see that the BNN-informed sampler reaches the maximal posterior value of the uninformed sampler already after about $3500$ samples (green vertical line). Overall this results in a better image interpretation (Figure \ref{fig:inf-informed}) compared to the uninformed sampler (Figure \ref{fig:inf-uninformed}). 

We test the significance of our result by evaluating the informed and uninformed samplers over a population of 150 face images from the CMU-Multipie dataset. We compare the maximal posterior observed with a ranked Friedmann test and obtain a p-value of $5.7\times10^{-5}$. This result highlights the superiority of our approach over an uninformed sampler in terms of exploring the maxima of the posterior within a fixed frame of samples.

\section{Discussion}
We have presented a novel approach to inform MCMC sampling with Bayesian Neural Networks. In our experiments we demonstrate that:

\textbf{BNNs allow for the estimation of an image-dependent proposal distribution.} Our qualitative results indicate that the BNN estimate is a meaningful measure of the uncertainty in the 3D face reconstruction process (Figure \ref{fig:bnn}).

\textbf{BNN-Informed MCMC significantly improves the exploration of maximal posterior regions} compared to an uninformed Gaussian random walk. An extensive evaluation of our approach on a population of 150 face images demonstrated a highly significant improvement in terms of the observed the face reconstruction quality (Figure \ref{fig:inf}).


\section*{Acknowledgment} A.K. is  supported  by  a  Novartis  University of Basel Excellence Scholarship for Life Sciences. M.W., A.W. and S.P are partially supported by the Swiss National Science Foundation (SNF), SystemsX.ch and the National Center of Competence in Research MARVEL. We gratefully acknowledge the support of NVIDIA with the donation of a Titan Xp.
\small{
    \bibliography{paper}
    \bibliographystyle{plain}
}

\end{document}